\documentclass[10pt,twocolumn,letterpaper]{article}

\usepackage[pagenumbers]{cvpr} 

\usepackage{lineno}

\definecolor{cvprblue}{rgb}{0.21,0.49,0.74}
\usepackage[pagebackref,breaklinks,colorlinks,allcolors=cvprblue]{hyperref}

\def\MethodName{FloVD}
\usepackage{kotex}
\usepackage[accsupp]{axessibility}
\usepackage[capitalize]{cleveref}
\Crefname{section}{Sec.}{Secs.}
\Crefname{section}{Section}{Sections}
\Crefname{table}{Table}{Tables}
\crefname{table}{Tab.}{Tabs.}
\usepackage[table]{colortbl}

\def\paperID{6295} 
\def\confName{CVPR}
\def\confYear{2025}

\title{FloVD: Optical Flow Meets Video Diffusion Model for \\ Enhanced Camera-Controlled Video Synthesis}

\author{Wonjoon Jin$^{1,2\dag}$
 \and Qi Dai$^{2}$ \and Chong Luo$^{2}$ \and Seung-Hwan Baek$^{1}$ \and Sunghyun Cho$^{1}$ \\
\vspace{-4.5mm}
\and
$^1$POSTECH \and $^2$Microsoft Research Asia \\
}

\begin{document}


\twocolumn[{%
\renewcommand\twocolumn[1][]{#1}%
\maketitle
\vspace{-1.5em}
\centering
\includegraphics[width=0.99\linewidth]{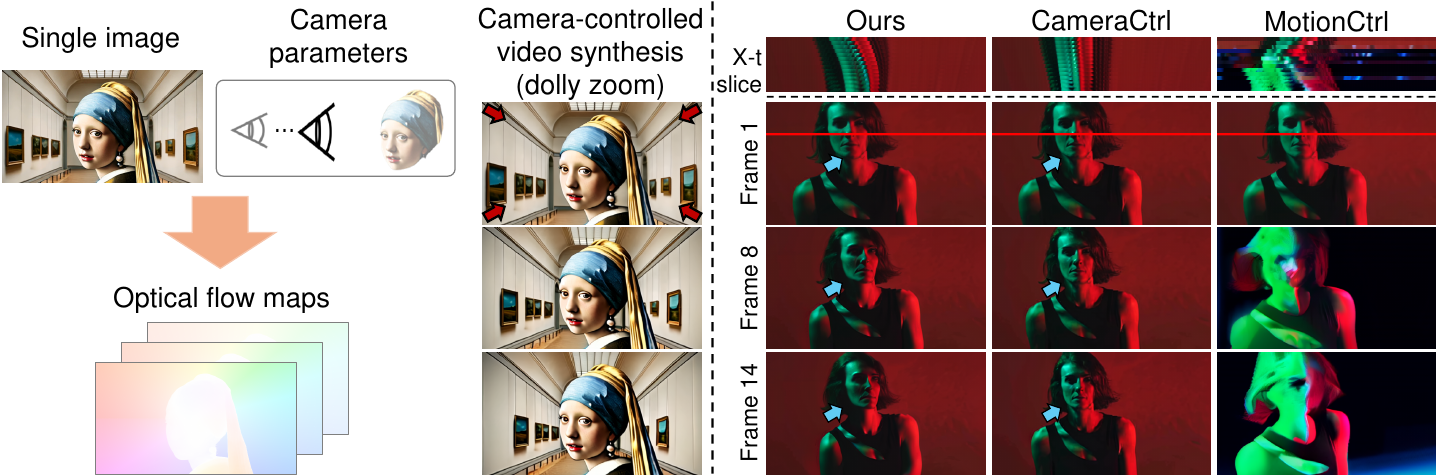}
\vspace{-3mm}
\captionof{figure}{
    (Left) Our method using optical flow enables video synthesis with complex camera movements (dolly zoom).
    (Right)
    Synthesized video frames with 'zoom-out' camera motion.
    X-t slice reveals pixel value changes along the red line.
    Our method shows natural object motion and accurate camera control, while CameraCtrl~\cite{he2024cameractrl} produces an object without motions, 
    and MotionCtrl~\cite{wang2024motionctrl} produces artifacts.
}
\vspace{3.5mm}
\label{fig:teaser}
}]

\vspace{-6.5mm}
{\let\thefootnote\relax\footnotetext{\noindent ${}^{\dag}$Work done during an internship at Microsoft Research Asia.}

\begin{abstract}

We present \MethodName{}, a novel video diffusion model for camera-controllable video generation.
\MethodName{} leverages optical flow to represent the motions of the camera and moving objects.
This approach offers two key benefits.
Since optical flow can be directly estimated from videos, our approach allows for the use of arbitrary training videos without ground-truth camera parameters.
Moreover, as background optical flow encodes 3D correlation across different viewpoints, our method enables detailed camera control by leveraging the background motion.
To synthesize natural object motion while supporting detailed camera control, our framework adopts a two-stage video synthesis pipeline consisting of optical flow generation and flow-conditioned video synthesis.
Extensive experiments demonstrate the superiority of our method over previous approaches in terms of accurate camera control and natural object motion synthesis.

\end{abstract}    
\vspace{-4.5mm}

\section{Introduction}
\label{sec:intro}

Video diffusion models have made significant strides in generating high-quality videos by leveraging large-scale datasets \cite{ho2022video,blattmann2023align,he2022latent,singer2022make,yang2024cogvideox,videoworldsimulators2024,blattmann2023stable,wang2024microcinema,polyak2024movie}. However, they often lack the ability to incorporate user-defined controls, particularly in terms of camera movement and perspective. This limitation restricts the practical applications of video diffusion models, where precise control over camera parameters is crucial for various tasks such as film production, virtual reality, and interactive simulations.

Recently, several approaches have introduced camera controllability to video diffusion models. One line of methods uses either text descriptions or user-drawn strokes that describe background motion as conditional inputs to represent camera motion~\cite{chen2023motion, yin2023dragnuwa, shi2024motion, wang2024motionctrl, polyak2024movie}. However, these methods support only limited camera controllability, such as zoom and pan, during video generation.

More sophisticated camera control has been achieved by directly using camera parameters as inputs~\cite{wang2024motionctrl, he2024cameractrl, xu2024camco, zheng2024cami2v, bahmani2024vd3d, yang2024direct, kuang2024collaborative,zhang2024recapture,xu2024cavia,wang2024humanvid}. 
In particular, recent methods embed input camera parameters using the Pl\"ucker embedding scheme~\cite{sitzmann2021light}, which involves embedding ray origins and directions, and feed them into video diffusion models~\cite{he2024cameractrl, xu2024camco, zheng2024cami2v, bahmani2024vd3d}.
While these approaches offer more detailed control, they require a training dataset that includes ground-truth camera parameters for every video frame. Acquiring such datasets is challenging, leading to the use of restricted datasets that primarily consist of static scenes, such as RealEstate10K~\cite{zhou2018stereo}. 
Consequently, they suffer from limited generalization capability, producing videos with unnatural object motions and inaccurate camera control (\cref{fig:teaser}).

To enable natural object motion synthesis and accurate camera control,
our key idea is to use \emph{optical flow} as conditional input to a video diffusion model.
This approach provides two key benefits.
First, since optical flow can be directly estimated from videos, our method eliminates the need for using datasets with ground-truth camera parameters.
This flexibility enables the utilization of arbitrary training videos.
Second, since background optical flow encodes 3D correlations across different viewpoints, our method enables detailed camera control by leveraging the background motion.
As a result, our approach facilitates natural object motion synthesis and precise camera control (\cref{fig:teaser}).

Based on the key idea, this paper presents \emph{\MethodName{}}, a novel camera-controllable video generation framework that leverages optical flow.
Given a single image and camera parameters, \MethodName{} synthesizes future frames with a desired camera trajectory.
To this end, our framework employs a two-stage video synthesis pipeline: optical flow generation and flow-conditioned video synthesis.
We generate optical flow maps representing the motions of the camera and moving objects from the input image and camera parameters.
These flow maps are then fed into a flow-conditioned video synthesis model to generate the final output video.

To synthesize natural object motion while supporting detailed camera control, \MethodName{} divides the optical flow generation stage into two sub-problems: camera flow generation and object flow generation.
First, we convert input camera parameters into optical flow of the background motion by using 3D structure estimated from the input image.
Next, an object motion synthesis model is introduced to generate the optical flow for object motions based on the input image.
We obtain the final optical flow maps by combining the background and object motion flows.


Our contributions are summarized as follows:
\begin{itemize}
    \item We present a novel camera-controllable video generation framework that leverages optical flow, allowing our method to utilize arbitrary training videos without ground-truth camera parameters.
    \item To achieve detailed camera control and high-quality video synthesis, we adopt a two-stage video synthesis pipeline, flow generation and flow-conditioned video synthesis.
    \item Extensive evaluation demonstrates the effectiveness of our method, showcasing its ability to produce high-quality videos with 
    accurate camera control and natural object motion.
\end{itemize}
\section{Related Work}
\label{sec:related}

\paragraph{Camera-controllable video synthesis.}
Following the tremendous success of video diffusion models, numerous efforts have been made to integrate camera controllability into the video generation process.
MovieGen~\cite{polyak2024movie} uses text descriptions that describe camera motions to control camera motion.
MCDiff~\cite{chen2023motion}, DragNUWA~\cite{yin2023dragnuwa}, and MotionI2V~\cite{shi2024motion} enable camera control through user-provided strokes, manipulating background motion to adjust camera movement.
AnimateDiff~\cite{guo2023animatediff} and Direct-a-video~\cite{yang2024direct} enable camera control by training models on augmented video datasets that contain simple camera movements, such as translation.
While these methods offer basic camera control with high-level instructions such as zoom and pan, they lack the detailed control capability for user-defined specific camera motions.

Recent methods have demonstrated detailed control of camera movement by using desired camera parameters as conditional input.
To this end, these approaches train models on datasets that provide ground-truth camera parameters for every video frame.
MotionCtrl~\cite{wang2024motionctrl} directly projects the camera extrinsic parameters onto the intermediate features of a diffusion model, while CameraCtrl~\cite{he2024cameractrl} leverages the Pl\"{u}cker embedding scheme~\cite{sitzmann2021light} to encode the camera origin and ray directions as conditioning input.
CamCo~\cite{xu2024camco} and CamI2V~\cite{zheng2024cami2v} enhance camera control through an epipolar attention mechanism across video frames.
VD3D~\cite{bahmani2024vd3d} enables camera motion control within the video synthesis process of transformer-based video diffusion models.

While these methods support detailed camera control, they are primarily trained on restricted datasets~\cite{zhou2018stereo} due to the requirement for camera parameters during training.
This limitation degrades the generalization capability and leads to unnatural object motion synthesis.
In contrast, our model can be trained on arbitrary videos by leveraging optical flow maps as input, which can be robustly estimated using recent optical flow estimation models~\cite{teed2020raft}.

\begin{figure*}[t!]
\centering
\includegraphics[width=0.97\linewidth]{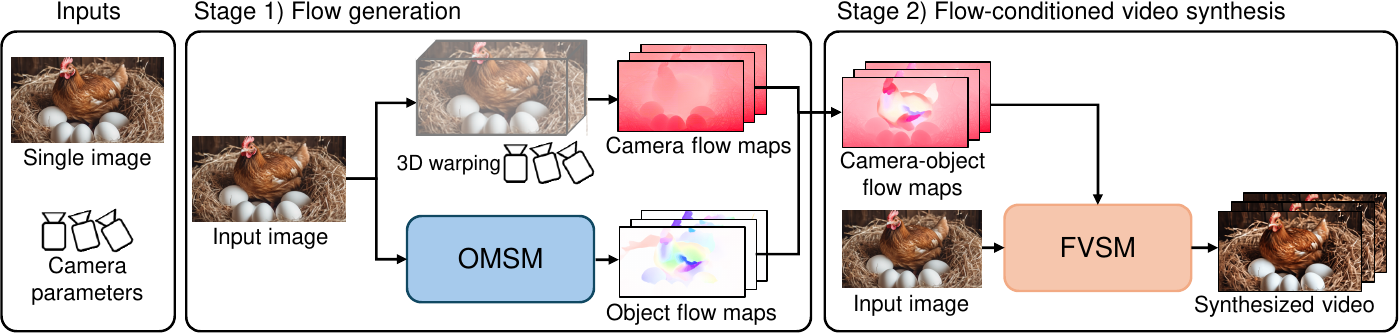}
\vspace{-3mm}
\caption{
Overview of \MethodName{}.
Given an image and camera parameters, our framework synthesizes video frames following the input camera trajectory.
To this end, we synthesize two sets of optical flow maps that represent camera and object motions. Then, two optical flow maps are integrated and fed into the flow-conditioned video synthesis model, enabling camera-controllable video generation.
}
\vspace{-3mm}
\label{fig:framework_overview}
\end{figure*}

\paragraph{Flow-based two-stage video synthesis.}
Recently, several studies have introduced optical-flow-based two-stage pipelines for video synthesis~\cite{li2024generative, holynski2021animating, endo2019animating, mahapatra2022controllable, zhao2022thin, ni2023conditional, liang2024movideo}.
These approaches, similar to ours, utilize two distinct models: one to generate optical flow maps and another to produce video frames based on the generated optical flow.
However, these approaches aim to improve video synthesis quality in terms of temporal coherence, and do not address camera-controlled video synthesis.
Furthermore, they do not distinguish between camera and object motions, thus extending them to incorporate camera control is not straightforward.

\section{\MethodName{} Framework}
\label{sec:method}
\cref{fig:framework_overview} presents an overview of \MethodName{}.
Our framework takes an image $I_s$ and camera parameters $\mathcal{C}=\{C_t\}_{t=1}^T$ as input where $t$ is a video frame index, and $T$ is the number of video frames.
$C_t$ is defined as a set of extrinsic and intrinsic camera parameters.
Given the input conditions, our framework synthesizes a video $\mathcal{I}=\{I_t\}^{T}_{t=1}$ that starts from $I_s$ as the first frame and follows the input camera trajectory, where $I_t$ is the $t$-th video frame.

\MethodName{} consists of two stages. First, the flow generation stage synthesizes two sets of optical-flow maps that represent camera and object motions using 3D warping and an object motion synthesis model (OMSM), respectively. We refer to these optical flow maps as camera flow maps and object flow maps.
These flow maps are integrated to form a single set of optical flow maps, which we refer to as camera-object flow maps. 
In the subsequent stage, a flow-conditioned video synthesis model (FVSM) synthesizes a video using the input image $I_s$ and the camera-object flow maps. In the following, we describe each stage in detail.

\subsection{Flow Generation}
\label{sec:flow_generation}
\paragraph{Camera flow generation.}
In the flow generation stage, we first generate camera flow maps $\mathcal{F}^c=\{f^c_t\}_{t=1}^{T}$ where $f^c_t$ is an optical flow map from the first frame to the $t$-th frame.
To generate camera flow maps reflecting the 3D structure in the input image $I_s$, we estimate a depth map $d_s$ from $I_s$ using an off-the-shelf single-image 3D estimation network~\cite{yang2024depth}.
Using the estimated depth map, we unproject each pixel coordinate $x_s$ in the input image $I_s$ into the 3D space.
Then, for each $t$, we warp the unprojected coordinates and project them back to the 2D plane using $C_t$ to obtain the warped coordinate $x_t$.
Finally, we construct the camera flow map $f_t^c$ by computing displacement vectors from $x_s$ to $x_t$.

\begin{figure}[t!]
\centering
\includegraphics[width=0.89\columnwidth]{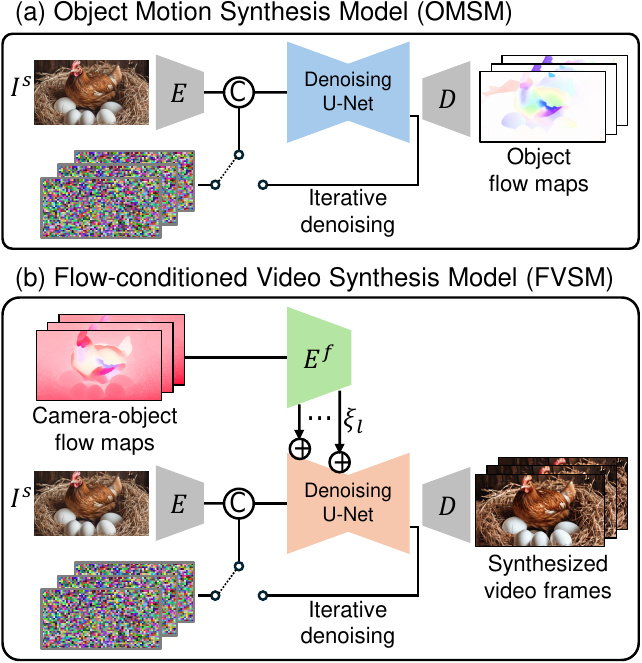}
\vspace{-3mm}
\caption{
Network architectures of OMSM and FVSM.
}
\vspace{-4mm}
\label{fig:diffusion_models}
\end{figure}

\paragraph{Object flow synthesis.}
In this stage, we also generate object flow maps $\mathcal{F}^o=\{f_t^o\}_{t=1}^T$ that represent object motions independent of background motions.
To this end, we develop OMSM based on the latent video diffusion model~\cite{blattmann2023stable}.
Specifically, as shown in \cref{fig:diffusion_models}(a), OMSM consists of a denoising U-Net, and an encoder and decoder of the latent video diffusion model's variational autoencoder (VAE).
In OMSM, the input image $I_s$ is first encoded by the VAE encoder.
Then, the denoising U-Net takes a concatenation of the encoded input image and a noisy latent feature volume as input, and iteratively denoises the latent feature volume to synthesize latent object motion flow maps.
Finally, the VAE decoder decodes the synthesized result and produces object flow maps $\mathcal{F}^o$. 

Inspired by Marigold \cite{ke2024repurposing}, we utilize the VAE decoder of the latent video diffusion model, which is trained on RGB images, for decoding object flow maps without any architectural changes or fine-tuning. Specifically, from the three output channels of the VAE decoder corresponding to RGB, we use only the first two channels for the $x$ and $y$ components of an object flow map.
Our training process also involves the VAE encoder.
Like the VAE decoder, we use the VAE encoder of the latent video diffusion model without any architectural changes or fine-tuning. For the three input channels of the VAE encoder, we feed the $x$ and $y$ components of an object flow map, along with their average $(x+y)/2$.
We verified that the optical flow map can be reconstructed from the encoded latent feature with a negligible error without any modification of the VAE.

\paragraph{Flow integration.}
Once the camera and object flow maps, $\mathcal{F}^c$ and $\mathcal{F}^o$, are generated, we obtain camera-object flow maps $\mathcal{F}=\{f_t\}_{t=1}^T$ by combining them.
The integration is performed as follows.

First, we estimate a binary mask $M^{obj}$ from the input image $I_s$ using an off-the-shelf segmentation model \cite{ravi2024sam}, which indicates pixels corresponding to moving objects. We use a single binary mask $M^{obj}$ for $t$, as all flow maps are forward-directional optical flow maps from the first frame to the $t$-th frame. Based on $M^{obj}$, we combine $f_t^c$ and $f_t^o$.
Specifically, for each pixel $x$ specified by $M^{obj}$, we compute its displaced position $x'$ using the object motion in $f_t^o$ as $x' = x + f_{t,x}^o$, where $f_{t,x}^o$ is the optical flow vector in $f_t^o$ at pixel $x$. Next, we transform $x'$ using the camera parameter $C_t$ and the depth map $d_s$, obtaining $x'_t$, which represents the displaced position of $x$ at the $t$-th frame due to both camera and object motions. We then compute the flow vector $f'_{t,x} = x'_t - x$. 
Finally, we derive the camera-object flow map $f_t$ as:
\begin{equation}
f_{t,x} = (1 - M^{obj}_x) \cdot f_{t,x}^c + M^{obj}_x \cdot f'_{t,x},
\label{eq:flow_integration}
\end{equation}
where $M^{obj}_x$ is the binary value of $M^{obj}$ at pixel $x$.

It is important to note that physically valid integration of camera and object flows requires object motion information along the $z$-axis (orthogonal to the image plane), which is not captured in the object flow maps.
Thus, our integration process does not produce physically accurate camera-motion flow maps.
However, we experimentally found that our framework can still synthesize videos with natural object motions.
This is made possible by the flow-conditioned video synthesis model (FVSM), which is trained on natural-looking videos, ensuring realistic object motions, even for input noisy camera-motion flow maps.

Our OMSM is trained to generate non-zero flow vectors only for dynamic objects. Therefore, we do not necessarily need to use the mask $M^{obj}$ in \cref{eq:flow_integration}; instead, we can transform the entire object motion flow maps using the camera parameters. However, we empirically found that using the mask $M^{obj}$ improves video synthesis quality by removing incorrectly synthesized flow vectors in static regions of $f_t^o$.

\subsection{Flow-Conditioned Video Synthesis}
\label{sec:video_synthesis_model}

The flow-conditioned video synthesis stage synthesizes a video $\mathcal{I}$ using the input image $I_s$ and the camera-object flow maps $\mathcal{F}$ as conditions.
To achieve this, our framework utilizes FVSM, which extends the latent video diffusion model~\cite{blattmann2023stable} by incorporating an additional flow encoder inspired by the T2I-Adapter architecture~\cite{mou2024t2i}.
Specifically, as shown in \cref{fig:diffusion_models}(b), our model consists of a flow encoder $E^f$, a denoising U-Net, and a VAE encoder and decoder.

The flow encoder takes the input camera-object flow maps $\mathcal{F}$, and computes multi-level flow embeddings:
\begin{equation}
\begin{aligned}
\{\xi_{(t,l)}\}^{T,L}_{t=1,l=1}=E^f(\mathcal{F}),
\end{aligned}
\end{equation}
where $\xi_{(t,l)}$ is a flow embedding of the $t$-th frame $I_t$ at level $l$. Each flow embedding has the resolution of its corresponding layer's latent feature in the denoising U-Net.
The denoising U-Net takes a concatenation of the encoded input image and a noisy latent feature volume as input, and iteratively denoises the latent feature volume of the video.
Additionally, the denoising U-Net also takes the multi-level flow embeddings by adding each of them to the feature at each layer of the denoising U-Net.
Finally, the synthesized video frames are obtained by decoding the denoised latent feature volume using the VAE decoder.
More details on the network architecture can be found in the supplemental document.

\section{\MethodName{} Training}
\label{sec:Training}

\MethodName{} utilizes two diffusion models: OMSM and FVSM, which are trained separately. As discussed in \cref{sec:intro}, both models can be effectively trained using a wide range of videos with dynamic object motions without requiring ground-truth camera parameters, thanks to our optical-flow-based representation.
In the following, we explain our datasets and training strategies for our models.

\subsection{Training Datasets}
For training these diffusion models, we primarily use an internal dataset containing 500K video clips, and its subset of video clips without camera motions. We refer to these as the full dataset and the curated dataset, respectively. The full dataset contains scenes similar to those in the Pexels dataset~\cite{pexels_dataset}.
The curated dataset contains around 100K video clips.
For training OMSM, we use both datasets, while for training FVSM, we use only the full dataset.

Training the diffusion models in our framework requires optical flow maps for each video clip.
We estimate the optical flow maps using an off-the-shelf estimator~\cite{teed2020raft}, and use them as the ground-truth object flow maps for OMSM, and the camera-object flow maps for FVSM.

The curated dataset is generated through the following process. For each video clip in the full dataset, we first detect the static background region from the first frame using an off-the-shelf semantic segmentation model~\cite{ravi2024sam}. Next, we compute the average magnitude of the optical flow vectors for all video frames within the background region. If this average magnitude is smaller than a specified threshold, we consider the video clip to have no camera motion and include it in the curated dataset.

\subsection{Training Object Motion Synthesis}
\label{sec:Training_MSM}
OMSM is trained in two stages.
The first stage initializes the model with the parameters of a pre-trained video diffusion model, and trains the model on the full dataset.
The second stage fine-tunes the model using the curated dataset without camera motions.
During training, we only update the parameters of the denoising U-Net, while fixing the parameters of the VAE encoder and decoder.
We train the model via denoising score matching~\cite{karras2022elucidating}.

The two-stage approach helps overcome the domain difference between the video synthesis task of the pretrained model and the object motion synthesis task, allowing for effective learning of object motion synthesis from the small-scale curated dataset.
\cref{fig:Data_Curation} shows an example of synthesized motions after the first and second training stages.
After the first stage, OMSM is effectively trained to synthesize object motion flow maps, but they exhibit camera motions in the background.
After the second stage, the model can successfully synthesize object motion flow maps with minimal camera motions.

\begin{figure}[t!]
\centering
\includegraphics[width=\columnwidth]{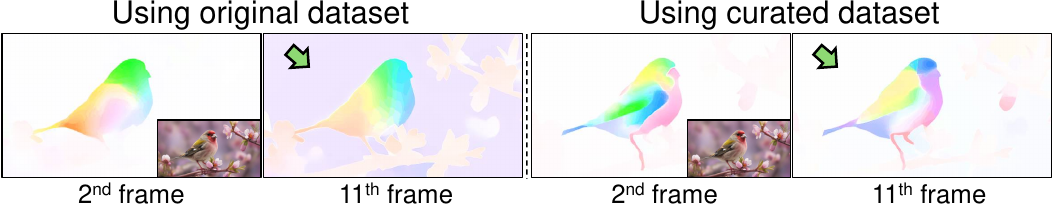}
\vspace{-6.3mm}
\caption{
Object flow maps synthesized by OMSM, which is trained on the full dataset (left) and the curated dataset (right), respectively.
White indicates optical flow vectors with no motion.
}
\vspace{-5mm}
\label{fig:Data_Curation}
\end{figure}

\subsection{Training Flow-Conditioned Video Synthesis}
\label{sec:Training_VSM}
We initialize FVSM using the parameters of a pretrained video diffusion model~\cite{blattmann2023stable}.
Then, we train only the flow encoder while fixing the other components.
Similar to OMSM, we train FVSM via denoising score matching~\cite{karras2022elucidating}.
While the optical flow maps are directly estimated from video datasets in the training time, the camera-object flow maps used in the inference time are synthesized through 3D warping and OMSM. Nevertheless, both optical flow maps contain camera and object motions in the form of flow vectors, enabling the FVSM to produce natural videos with the desired camera motion effectively.

\section{Experiments}
\label{sec:experiment}

\subsection{Implementation Details}
\label{sec:implementation_detail}
\MethodName{} synthesizes 14 video frames at once, following Stable Video Diffusion~\cite{blattmann2023stable}.
We use a resolution of $320 \times 576$ for both video frames and optical flow maps.
FVSM is trained for 50K iterations with 16 video clips and their optical flow maps per training batch.
OMSM is trained on the full dataset for 100K iterations and then fine-tuned using the curated dataset for 50K iterations, with 8 optical flow maps per training batch.
Inspired by T2I-Adapter~\cite{mou2024t2i}, we use a quadratic timestep sampling strategy (QTS) in training FVSM for better camera controllability~(Tab.~S2 in the supplemental document).
For stable training and inference of \MethodName{}, we adaptively normalize optical flow maps based on statistics computed from the training dataset, following Li et al.~\cite{li2024generative}.
Refer to the supplemental document for more implementation details.

\begin{figure*}[!t]
\centering
\begin{tabular}{@{}c}
\includegraphics[width=0.95\linewidth]{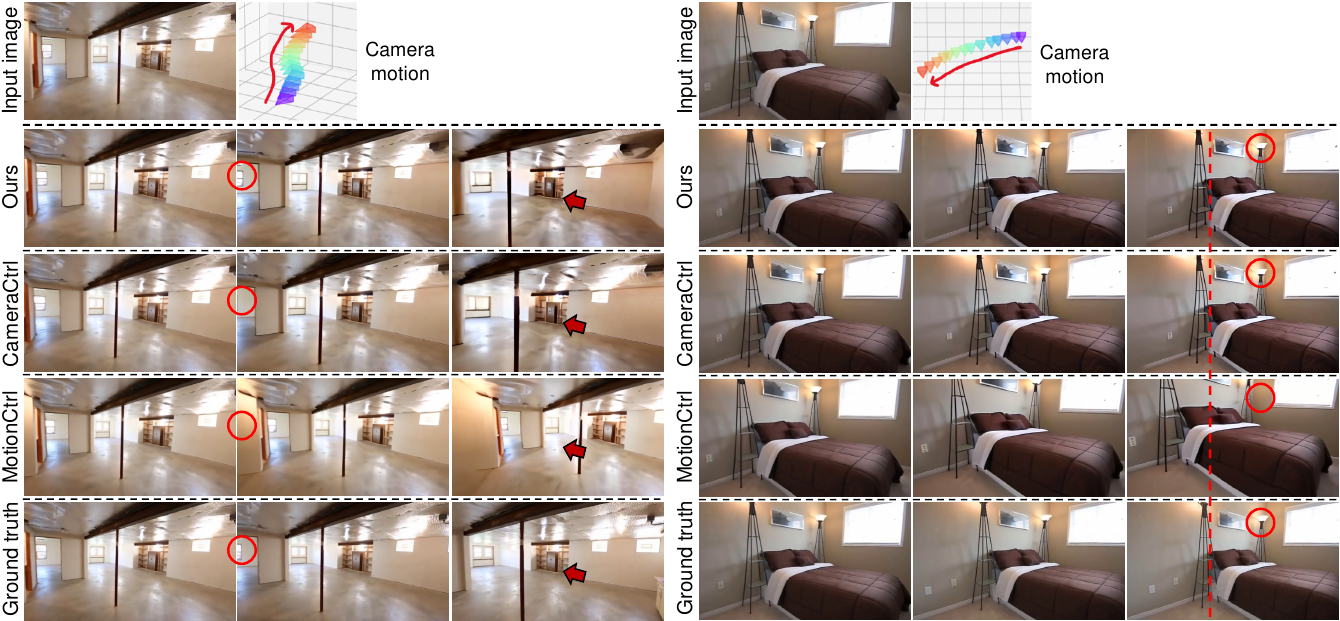} \\
\end{tabular}
\vspace{-4mm}
\caption{
Qualitative comparison of camera control using the RealEstate10K test dataset~\cite{zhou2018stereo}.
MotionCtrl~\cite{wang2024motionctrl} often fails to follow the input camera parameters.
Notably, our method shows accurate camera control results despite not using camera parameters in training.
}
\label{fig:qual_cam_control}
\vspace{-4.5mm}
\end{figure*}

\subsection{Evaluation Protocol}
\label{sec:evaluation_protocol}

\paragraph{Camera controllability.}
We evaluate the camera controllability following previous methods~\cite{he2024cameractrl,zheng2024cami2v}.
For an input image and camera parameters, we first synthesize a video.
We then estimate camera parameters from the synthesized video using GLOMAP~\cite{pan2024global}, and compare them against the input camera parameters to evaluate how faithfully the synthesized video follows the input parameters.
For the evaluation, we sampled 1,000 video clips and their associated camera parameters from the test set of RealEstate10K~\cite{zhou2018stereo}.

We employ the evaluation protocol of previous methods~\cite{he2024cameractrl,zheng2024cami2v} for the camera controllability. Specifically, for an input image and camera parameters, we first synthesize a video. We then estimate camera parameters from the synthesized video using GLOMAP~\cite{pan2024global}, and compare the estimated camera parameters against the input parameters to evaluate how faithfully the synthesized video follows the input camera parameters. For the evaluation dataset, we sampled 1,000 video clips and their associated camera parameters from the test set of RealEstate10K~\cite{zhou2018stereo}.

To evaluate estimated camera parameters against input ones,
we measure the mean rotation error (mRotErr), mean translation error (mTransErr), and mean error in camera extrinsic matrices (mCamMC), which are defined as:
\begin{equation}
\vspace{-2.5mm}
\begin{aligned}
\textnormal{mRotErr}&=\frac{1}{T} \sum\limits_{t=1}^{T} \cos^{-1}\frac{\textrm{tr}(\hat{R_t}R_t^{T}) - 1}{2}, \\
\vspace{-2.3mm}
\textnormal{mTransErr}&=\frac{1}{T} \sum\limits_{t=1}^{T} \lVert \hat{\tau}_t - \tau_{t} \rVert,~~~~~~\textrm{and} \\
\vspace{-2.3mm}
\textnormal{mCamMC}&=\frac{1}{T} \sum\limits_{t=1}^{T} \lVert [\hat{R}_t|\hat{\tau}_t] - [R_t|\tau_{t}] \rVert_{2}, \\
\end{aligned}
\end{equation}
where $T$ is the number of video frames.
$\hat{R}_t$ and $\hat{\tau}_t$ are the camera rotation matrix and translation vector estimated from the $t$-th synthesized video frame, and $R_t$ are $\tau_t$ are their corresponding input rotation matrix and translation vector, respectively.

\vspace{-3mm}
\paragraph{Video synthesis quality.}
We evaluate the video synthesis quality in terms of (1) sample quality and (2) object motion synthesis quality.
For sample quality, we first construct a benchmark dataset using 1,500 videos randomly sampled from the Pexels dataset~\cite{pexels_dataset} (Pexels-random).
For the model's capability of diverse object motion synthesis, we construct three benchmark video datasets with small, medium, and large object motions, each containing 500 video clips with minimal camera motions to avoid potential bias caused by camera motion.
The datasets are categorized based on the average magnitudes of the optical flow vectors of moving objects: smaller than 20 pixels (Pexels-small), between 20 and 40 pixels (Pexels-medium), and more than 40 pixels (Pexels-large).
More details on the benchmark datasets can be found in the supplemental document.

To evaluate the video synthesis quality, we synthesize videos using the first frames of videos in the aforementioned  benchmark datasets and compare these synthesized videos with the datasets.
While our method's video synthesis quality is minimally affected by these parameters, previous methods that synthesize video frames directly from them might be more influenced. 
To account for this, we utilize seven types of camera trajectories during video synthesis: translation to the left, right, up, and down, as well as zoom-in, zoom-out, and no camera motion (`stop').
Consequently, for all models, we generate seven videos for each video included in the benchmark datasets.
Finally, we evaluate the video synthesis performance of a given method through the Frechet Video Distance (FVD)~\cite{unterthiner2018towards}, Frechet Image Distance (FID)~\cite{heusel2017gans}, and Inception Score (IS)~\cite{salimans2016improved}.

\begin{table}[!t]
\centering
\resizebox{\columnwidth}{!}{
    \begin{tabular}{l|c|ccc}
    \toprule[1pt]
                    & Training Data     & mRotErr ($^\circ$) & mTransErr        & mCamMC     \\ \hline
    MotionCtrl      & RE10K             & 5.90              & 0.1610            & 0.1719    \\
    CameraCtrl      & RE10K             & 1.44              & 0.0927            & 0.0945    \\ 
    Ours            & RE10K             & \textbf{1.43}     & \textbf{0.0869}   & \textbf{0.0887}    \\
    Ours            & Internal          & 1.79              & 0.0994            & 0.1018   \\
    Ours w/ OMSM    & RE10K             & 1.52              & 0.0971            & 0.0989    \\
    Ours w/ OMSM    & Internal          & 1.88              & 0.1042            & 0.1066    \\
         
    \bottomrule[1pt]
    \end{tabular}
}
\vspace{-3.0mm}
\caption{
Quantitative evaluation of camera controllability using the RealEstate10K test dataset~\cite{zhou2018stereo}. 
Our method shows superior camera control performance against previous methods~\cite{he2024cameractrl, wang2024motionctrl}, even without using camera parameters in training.
}
\vspace{-4.0mm}
\label{table:cam_control}
\end{table}

\subsection{Comparison}

We compare our method with recent camera-controllable video synthesis methods, MotionCtrl~\cite{wang2024motionctrl} and CameraCtrl~\cite{he2024cameractrl}, both of which support detailed camera control by taking camera parameters as input.
Additional comparisons with other methods that support basic camera movements can be found in the supplemental document.

\begin{figure*}[!t]
\centering
\begin{tabular}{@{}c}
\includegraphics[width=0.95\linewidth]{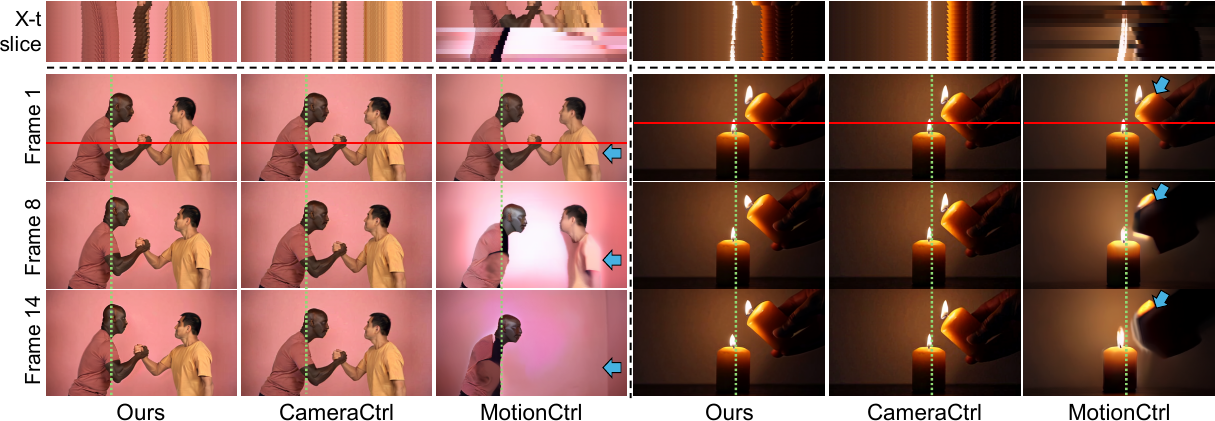} \\
\end{tabular}
\vspace{-4mm}
\caption{
Qualitative comparison of video synthesis quality.
Video frames are synthesized with 'stop' camera motion.
X-t slice reveals how pixel value changes over time along the horizontal red line.
MotionCtrl~\cite{wang2024motionctrl} often fails to follow input camera trajectory and synthesizes video frames with artifacts, due to the lack of generalization capability.
CameraCtrl~\cite{he2024cameractrl} frequently synthesizes motionless object in generated videos.
Our method synthesizes video frames with natural object motion while supporting precise camera control.
}
\label{fig:qual_obj_motion}
\vspace{-3mm}
\end{figure*}

\vspace{-4mm}
\paragraph{Camera controllability.}
We first compare the camera controllability of our method against MotionCtrl~\cite{wang2024motionctrl} and CameraCtrl~\cite{he2024cameractrl}.
Both MotionCtrl and CameraCtrl were trained on RealEstate10K~\cite{zhou2018stereo}, which provides no object motions but a wider range of camera motions than our full dataset.
For a comprehensive comparison, we evaluate four versions of our model.
Specifically, we train FVSM on either our internal dataset or RealEstate10K, but without utilizing the ground-truth camera parameters available in RealEstate10K.
We also include variants of our model with and without OMSM, as RealEstate10K contains only static scenes without moving objects.
In this evaluation, OMSM is trained using our internal dataset.

As shown in \cref{fig:qual_cam_control}, 
MotionCtrl~\cite{wang2024motionctrl} produces video frames that do not accurately follow the input camera trajectories due to its suboptimal camera parameter embedding scheme.
On the other hand, both CameraCtrl~\cite{he2024cameractrl} and ours accurately reflect the input camera parameters, and produce video frames that closely resemble the ground-truth frames.

As reported in \cref{table:cam_control}, our model trained on RealEstate10K~\cite{zhou2018stereo} outperforms both MotionCtrl and CameraCtrl across all metrics.
Moreover, our other models show comparable performances to CameraCtrl, while using the internal dataset and incorporating OMSM slightly increase errors due to domain differences and object motions.
These results prove the effectiveness of our camera control scheme based on optical flow.

\vspace{-4.5mm}
\paragraph{Video synthesis quality.}
We also compare the video synthesis quality of our method with previous ones~\cite{he2024cameractrl,wang2024motionctrl}.
\cref{fig:qual_obj_motion} shows a qualitative comparison, including X-t slices to visualize pixel value changes over time, computed from the positions marked by the red lines.
In this comparison, we synthesize videos using camera parameters without any motion, mainly to compare the video synthesis quality.
CameraCtrl~\cite{he2024cameractrl} produces results with no object motions, as shown in its X-t slices.
MotionCtrl~\cite{wang2024motionctrl} produces artifacts with inconsistent foreground and background regions, as marked by blue arrows.
These artifacts result from the limited generalization capability, since MotionCtrl updates certain pre-trained parameters of the video diffusion model during training.
Unlike these methods, our method produces high-quality videos with natural object motions.

The superior performance of our method is also evidenced by the quantitative comparison in \cref{table:sample_quality}.
For the Pexels-random dataset, our method reports better sample quality against the previous methods~\cite{he2024cameractrl,wang2024motionctrl}.
These results prove that our method does not harm the video synthesis quality of the pre-trained video diffusion model, compared to the previous ones.

Our method also achieves better performances for the benchmark datasets of object motion synthesis quality (Pexels-small, Pexels-medium, and Pexels-large), as reported in \cref{table:sample_quality}.
While CameraCtrl exhibits significantly degraded quality for large object motions (Pexels-large), our method achieves substantially better results for all three benchmark datasets.
MotionCtrl often fails to follow input camera parameters, synthesizing videos where the viewpoint remains close to the input image. This may lead to good FVD scores, as the synthesized videos align well with the minimal camera movement present in most benchmark videos.
However, as shown in \cref{fig:qual_obj_motion}, MotionCtrl often produces visual artifacts in the synthesized videos.
These artifacts are also evidenced by the degraded FID scores for MotionCtrl.
More visual examples of the visual artifacts can be found in the supplemental document.
In addition, by employing a timestep sampling strategy from the EDM framework~\cite{karras2022elucidating}, our method outperforms previous methods across all metrics (Tab.~S1 in the supplemental document).

\begin{table}[!t]
\centering
\resizebox{\linewidth}{!}{
    \setlength\tabcolsep{1.5pt}
    \begin{tabular}{l|ccc|ccc|ccc|ccc}
    \toprule[1pt]
                    & \multicolumn{3}{c|}{Pexels-random} & \multicolumn{3}{c|}{Pexels-small ($<20$)} & \multicolumn{3}{c|}{Pexels-med. ($<40$)} & \multicolumn{3}{c}{Pexels-large ($\ge40$)}  \\    
                    & FVD        & FID        & IS         & FVD             & FID         & IS           & FVD          & FID         & IS         & FVD             & FID        & IS        \\ \hline
    MotionCtrl      & 93.54          & 36.06      & 11.19  & 235.39      & 28.94      & 10.53      & 214.47          & 25.76      & 10.74       & \cellcolor{yellow!100}\textbf{188.89} & 25.84   & 11.71           \\
    CameraCtrl      & 151.06         & 40.69      & 11.23  & 226.40       & 25.57      & 10.63      & 328.71          & 24.54      & 10.76       & 340.36       & 25.53      & 11.68           \\ 
    Ours            & \cellcolor{yellow!100}\textbf{91.55} & \cellcolor{yellow!100}\textbf{35.66} & \cellcolor{yellow!100}\textbf{11.63}   & \cellcolor{yellow!100}\textbf{220.65}& \cellcolor{yellow!100}\textbf{22.49}  & \cellcolor{yellow!100}\textbf{11.58} & \cellcolor{yellow!100}\textbf{183.14}  & \cellcolor{yellow!100}\textbf{20.71}    & \cellcolor{yellow!100}\textbf{11.68}     & 207.39     & \cellcolor{yellow!100}\textbf{21.12} & \cellcolor{yellow!100}\textbf{12.95}  \\       
    \bottomrule[1pt]
    \end{tabular}
}
\vspace{-3mm}
\caption{
Quantitative evaluation of video synthesis quality using the Pexels dataset~\cite{pexels_dataset}.
Our method shows superior video synthesis performance against previous methods~\cite{wang2024motionctrl,he2024cameractrl}.
}
\vspace{-3mm}
\label{table:sample_quality}
\end{table}

\subsection{Analysis}
\label{sec:ablation_study}
In the following, we provide an ablation study of our main components and a detailed mechanism of FVSM. 
Refer to Sec.~5 in the supplemental document for further analysis.

\vspace{-3mm}
\paragraph{Ablation study.}
We conduct an ablation study to verify the effect of our main components: OMSM, and training with a wide range of real-world videos, both of which are made possible by our optical-flow-based framework.
We utilize our models trained with two different timestep sampling strategies for in-depth analysis under different settings.
In \cref{fig:ablation_main}, the baseline model indicates a variant of our model that has no OMSM (i.e., only FVSM) and is trained on RealEstate10K~\cite{zhou2018stereo}.
`+~OMSM' indicates a variant model with OMSM, while its FVSM is still trained on RealEstate10K.
OMSM is trained using our full and curated datasets.
`+~large-scale data' is our final model where both OMSM and FVSM are trained using our datasets.

As shown in \cref{fig:ablation_main}(a), the baseline model does not synthesize noticeable object motions.
Introducing OMSM enables our framework to generate object motion, but it also occasionally produces artifacts for moving objects as shown in \cref{fig:ablation_main}(b).
Our final model produces natural-looking object motions without noticeable artifacts (\cref{fig:ablation_main}(c)). \cref{table:ablation_object} also shows similar trends that introducing each component to our framework consistently improves evaluation metrics for Pexels-random.
Additional quantitative ablation study can be found in Tab.~S1 of the supplemental document.

\begin{figure}[!t]
\centering
\includegraphics[width=0.93\linewidth]{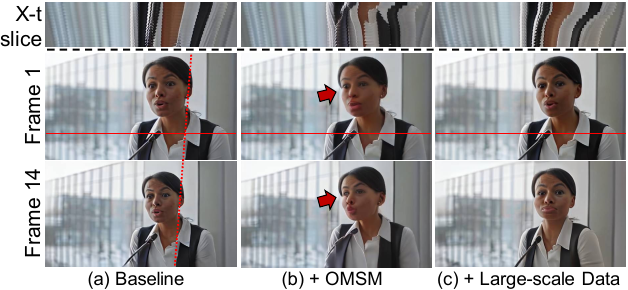}
\vspace{-3.0mm}
\caption{
Qualitative ablation with 'zoom-out' camera motion. X-t slice reveals pixel value changes along the horizontal red line. 
}
\vspace{-3.5mm}
\label{fig:ablation_main}
\end{figure}

\begin{table}[!t]
\centering
\resizebox{0.85\linewidth}{!}{
    \begin{tabular}{l|c|ccc}
    \toprule[1pt]
                    & Training  & \multicolumn{3}{c}{Pexels-random}  \\
                    & Data      & FVD ($\downarrow$)    & FID ($\downarrow$)     & IS ($\uparrow$)            \\ \hline
    Baseline    & RE10K  & 157.99   & 39.61  & 11.19       \\
    \hspace{1mm} + OMSM & RE10K & 104.92 & 36.33 & 11.51   \\       
    \hspace{1mm} + large-scale data  & Internal  & \textbf{91.55}  & \textbf{35.66} & \textbf{11.63}   \\ 
    \bottomrule[1pt]
    \end{tabular}
}
\vspace{-3mm}
\caption{
Ablation study of our main components with the evaluation of video synthesis quality using the Pexels-random dataset.
}
\vspace{-1mm}
\label{table:ablation_object}
\end{table}

\vspace{-3.5mm}
\paragraph{Flow-conditioned video synthesis.}
Our method generates camera flow maps using the 3D structure estimated from an input image and feeds them to FVSM. To better understand our framework, \cref{fig:explicit_control} presents visualizations of warped images using the estimated 3D structure and camera parameters, alongside their associated video synthesis results produced by FVSM.
As shown in the figure, 3D-based image warping may introduce distortions and holes, yet still provides realistic-looking images. This result indicates that leveraging the 3D structure can serve as a powerful hint for camera-controllable video synthesis. \cref{fig:explicit_control} also shows that our flow-conditioned video synthesis successfully produces realistic-looking results that closely resemble the warping results, but without artifacts such as distortions and holes.

\subsection{Applications}
\vspace{-1mm}
\paragraph{Temporally-consistent video editing.}
Our framework using FVSM enables temporally-consistent video editing at no extra cost.
Specifically, we first obtain optical flow maps from the input video and edit the first frame of the video.
We then synthesize a video by using FVSM with the edited first frame and flow maps as inputs, producing temporally-consistent video editing results~(\cref{fig:video_editing}).


\vspace{-4mm}
\paragraph{Cinematic camera control.}
Thanks to its 3D awareness, our framework supports advanced camera controls such as the dolly zoom, which moves the camera forward or backward while simultaneously adjusting the zoom in the opposite direction.
\cref{fig:teaser}(Left) shows a synthesized video with the dolly zoom effect, where the subject remains a similar size while the background appears to converge inward.
Notably, our framework accomplishes this without requiring training on video datasets with varying camera intrinsic parameters across frames.

\begin{figure}[!t]
\centering
\includegraphics[width=\columnwidth]{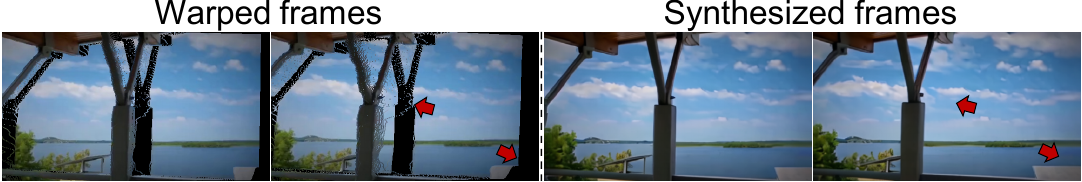}
\vspace{-6mm}
\caption{
Explicit camera control. Our model can follow the warped frames while handling artifacts such as holes, which are caused by imperfect 3D structure estimation.
}
\vspace{-2mm}
\label{fig:explicit_control}
\end{figure}

\begin{figure}[!t]
\centering
\includegraphics[width=0.9\linewidth]{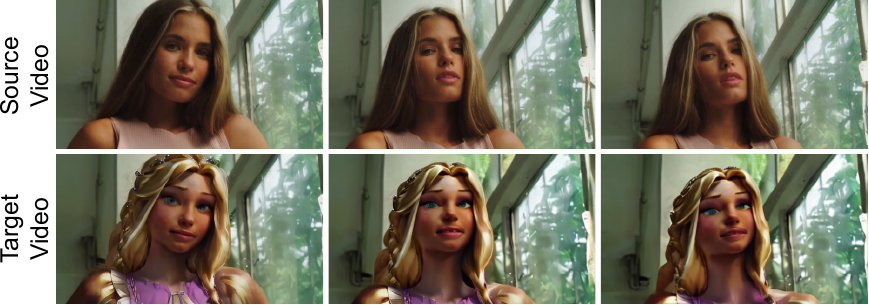}
\vspace{-3mm}
\caption{
Temporally-consistent video editing.
}
\label{fig:video_editing}
\vspace{-1mm}
\end{figure}

\section{Conclusion}
\label{sec:conclusion}
This paper proposed \MethodName{}, a novel optical-flow-based video diffusion model for camera-controllable video generation.
Since existing methods require a training dataset with ground-truth camera parameters, they are mainly trained on restricted datasets that primarily consist of static scenes, leading to video synthesis with unnatural object motion.
Unlike previous methods, our method leverages optical flow maps to represent both camera and object motions, enabling the use of arbitrary training videos without ground-truth camera parameters.
Moreover, our method facilitates detailed camera control by leveraging background motions of optical flow, which encodes 3D correlation across different viewpoints.
Our extensive experiments demonstrate that \MethodName{} provides realistic video synthesis with natural object motion and accurate camera control.

\vspace{-4mm}
\paragraph{Limitations.}
Our method is not free from limitations.
Errors from both the object motion synthesis model and the semantic segmentation model may result in unnatural object motion in the synthesized videos.
The estimation error of the segmentation model can be alleviated through user interaction by providing point prompts for object regions.
Our future work will involve a seamless integration of camera and object motions to synthesize more natural videos.

\vspace{-3.5mm}
\paragraph{Acknowledgment.}
This work was supported by the Korea government (MSIT), through the IITP grant (Global Research Support Program in the Digital Field program, RS-2024-00436680; Development of VFX creation and combination using generative AI, RS-2024-00395401; Artificial Intelligence Graduate School Program (POSTECH), RS-2019-II191906) and NRF grant (RS-2023-00211658; RS-2024-00438532), and Microsoft Research Asia.

\vspace{-3.5mm}
\paragraph{Ethical considerations.}
FloVD is purely a research project. Currently, we have no plans to incorporate FloVD into a product or expand access to the public. We will also put Microsoft AI principles into practice when further developing the models. In our research paper, we account for the ethical concerns associated with video generation research. To mitigate issues associated with training data, we have implemented a rigorous filtering process to purge our training data of inappropriate content, such as explicit imagery and offensive language, to minimize the likelihood of generating inappropriate content.

{
    \small
    \bibliographystyle{ieeenat_fullname}
    \bibliography{main}
}


\end{document}